\documentclass[runningheads]{llncs}
\usepackage{graphicx}
\usepackage{amsmath,amssymb} 
\usepackage{multirow}
\usepackage{color}
\usepackage{caption}
\usepackage{subfigure}
\usepackage[width=122mm,left=12mm,paperwidth=146mm,height=193mm,top=12mm,paperheight=217mm]{geometry}
\begin{document}
\pagestyle{headings}
\mainmatter
\def\ECCV18SubNumber{2082}  
\title{SparseNet: A Sparse DenseNet for Image Classification} 



\author{Wenqi Liu ~~~~~~~~~~~~~~~ Kun Zeng}

\institute{Sun Yat-sen University}
\maketitle

\begin{abstract}
Deep neural networks have made remarkable progresses on various computer vision tasks. Recent works have shown that depth, width and shortcut connections of networks are all vital to their performances. In this paper, we introduce a method to sparsify DenseNet which can reduce connections of a L-layer DenseNet from $O(L^2)$ to $O(L)$, and thus we can simultaneously increase depth, width and connections of neural networks in a more parameter-efficient and computation-efficient way. Moreover, an attention module is introduced to further boost our network's performance. We denote our network as $SparseNet$. We evaluate $SparseNet$ on datasets of CIFAR(including CIFAR10 and CIFAR100) and SVHN. Experiments show that $SparseNet$ can obtain
improvements over the state-of-the-art on CIFAR10 and SVHN. Furthermore, while achieving comparable performances as DenseNet on these datasets, $SparseNet$ is  $ \times 2.6$ smaller and $ \times 3.7$ faster than the original DenseNet.
\end{abstract}

\section{Introduction}

Deep convolutional neural networks have achieved great successes on many computer vision tasks, such as object classification, detection and segmentation
\cite{krizhevsky2012imagenet}
\cite{girshick2015fast}
\cite{long2015fully}.  `Depth' played a significant role while neural networks are achieving their successes. From AlexNet\cite{krizhevsky2012imagenet} to VGGNet\cite{simonyan2014very} and GoogLeNet\cite{szegedy2015going}, their performances on various computer vision tasks are boosting as network's depth is increasing.

Experiments\cite{he2016deep} have shown  if we simply stack layers without changing network's structure, its performance would get worse otherwise. Because gradients of network's parameters  will vanish  as depth is increasing.  To settle this problem, He\cite{he2016deep} proposed ResNet, which introduced a residual learning framework by adding identity-mapping shortcuts. ResNet extended its depth up to over 100 layers and achieved state-of-art performances in many computer vision tasks. However, when ResNet is  getting deeper(e.g. over 1000), it will suffer from the overfitting problem.

Huang\cite{huang2016deep} proposed a new  training procedure,  named stochastic depth, solved this problem. Take ResNet for example,  Huang\cite{huang2016deep} trained shallower subnetworks by randomly dropping residual modules(while retaining shortcut connections). The vanishing-gradient problem has been alleviated  since only shallower networks are trained in the training phase. This training procedure can extend depth of networks to over 1000 layers(e.g. 1202 layers) and the performance on image classification has been further improved.

Zagoruyko\cite{zagoruyko2016wide} improved ResNet from another aspect. He introduced a wider(more channels in convolution layers) and shallower ResNet variant. The performance of wide ResNet with only 16 layers exceeds that of original ResNet with over 1000 layers. Another benefit brought by wide ResNet is the training is very fast since it can take advantage of the parallel of GPU computing.

By gradually increasing width of neural networks, Han\cite{han2016deep} presented deep pyramidal residual Networks. For the original ResNet,  width only doubled after downsampling happened. For example, there are $4$ modules in original ResNet\cite{he2016deep}: $Conv2\_x$, $Conv3\_x$, $Conv4\_x$ and $Conv5\_x$.  Width for each module are $64$, $128$, $256$ and $512$. Within every module, dimensions are all the same. For pyramidal residual networks, width for each residual unit are always increasing no matter they are in the same module or not. Experiments shown pyramidal residual networks had superior generalization ability compared to the original residual networks. So except increasing the depth, properly increasing width is also another way to boost network's performance.

Besides increasing depth or width, increasing number of shortcut connections is another effective way of improving network's performance. It can gain network's performance from two aspects. 1)It shortens the distance between input and output and thus alleviates the vanishing-gradient problem with shorter forward flows. Highway networks\cite{srivastava2015highway} and ResNet\cite{he2016deep} proposed different ways of shortcut connections, both of which made training easier. 2)Shortcut connections can take advantage of multi-scale feature maps, which can improve performances on  various computer vision tasks\cite{hariharan2015hypercolumns}\cite{long2015fully}\cite{sermanet2013pedestrian}\cite{yang2015multi}.

Huang\cite{huang2017densely}takes this idea to the extreme. He proposed DenseNet, for the ${l^{th}}$ layer of which, it takes all previous ${(l-1)}$ layers as its input(connections of this layer is $O(l)$). By this kind of network structure design, it not only alleviates vanishing-gradient problem, but also achieves better feature reuse. DenseNet achieves superior performance on datasets of CIFAR-10,CIFAR-100 and SVHN. However, it has its own disadvantages. There are total ${\dfrac{L(L-1)}{2}}$ connections for a L-layer DenseNet. The excessive connections not only decrease networks' computation-efficiency and parameter-efficiency, but also make networks more prone to overfitting. As we can see from upper of Fig.1, When I modify connections of a 40-layer DenseNet, the test error rates on CIFAR10 first decrease and then increase as connection is increasing. When the connections is $22$, the error rate reaches the lowest, $5.11\%$.  However, as we can see from bottom of Fig.1, error rates on the training datasets is decreasing as connections is increasing.

\begin{figure}
\centering
\begin{minipage}[t]{0.5\linewidth}
\centering          
\includegraphics[scale=0.5]{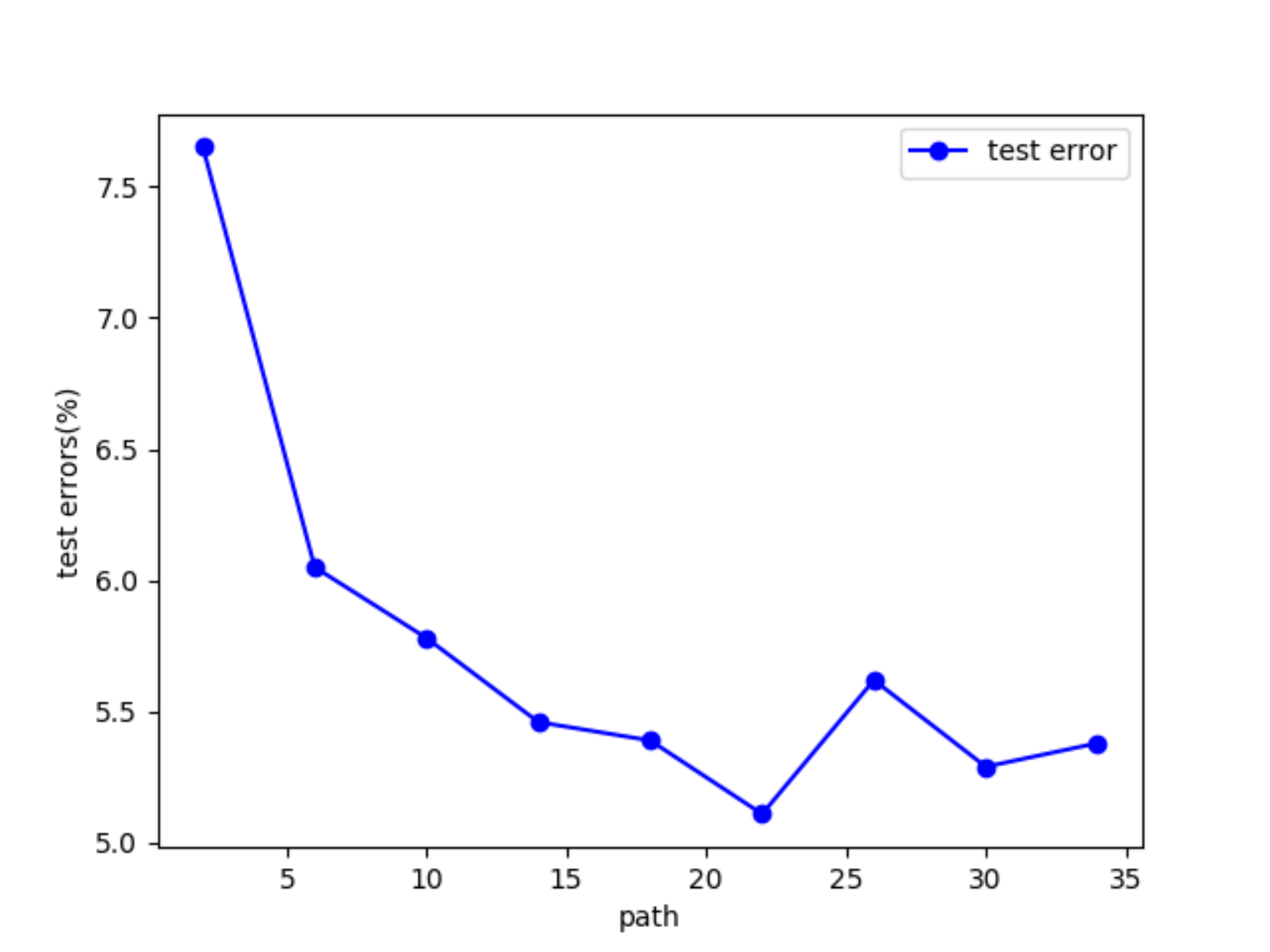}             
\end{minipage}

\begin{minipage}[t]{0.5\linewidth}
\centering  
\includegraphics[scale=0.5]{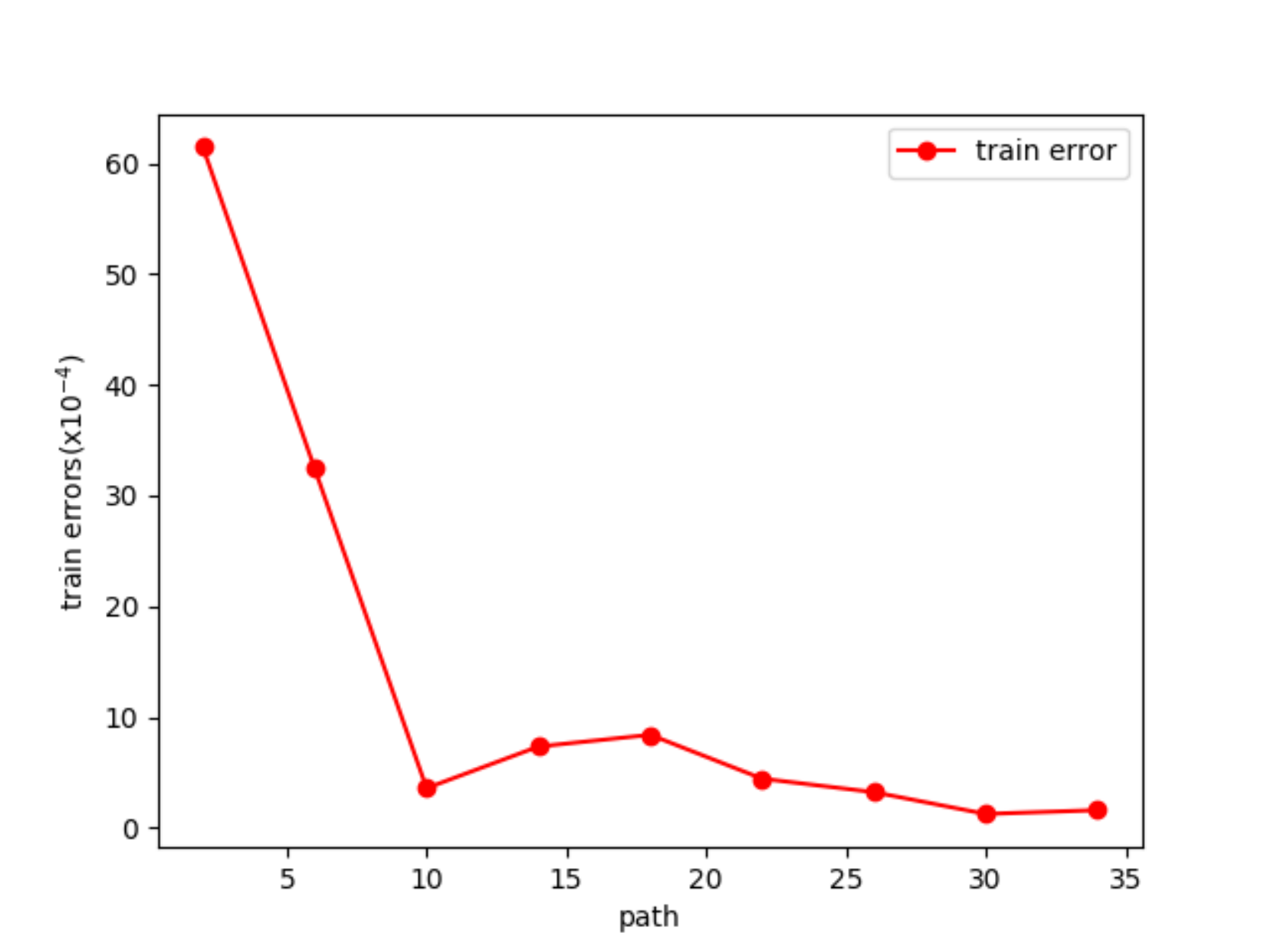}     
\end{minipage}

\caption{test/train error rate on CIFAR10 of different paths(connections) in DenseNet }
\label{fig:example}
\end{figure}

To settle this problem, we proposed a method to sparsify DenseNet. Zeiler\cite{zeiler2014visualizing} found out that for a deep neural network, shallower layers can learn concrete features, whereas deeper layers  can learn abstract features. Based on this observation, we can drop connections from middle layers and  reduce connections for each layer from O(n) to O(1). So  total connections of the sparsified DenseNet is $O(n)$.  As we can see in Fig. 2, left is a small part of DenseNet, right is a small part of SparseNet. the dotted line are dropped connections. So our idea for sparsifing is simply dropping connections from  middle layers and only retaining the nearest and farthest connections. And then we can extend the network to deeper or wider, which would result in better performance. As we can see in Fig. 2, while keeping the overall parameters unchanged, by dropping some connections and then extend network's width or depth, the performance of networks are getting better.

\begin{figure}
\centering
\includegraphics[height=6.5cm]{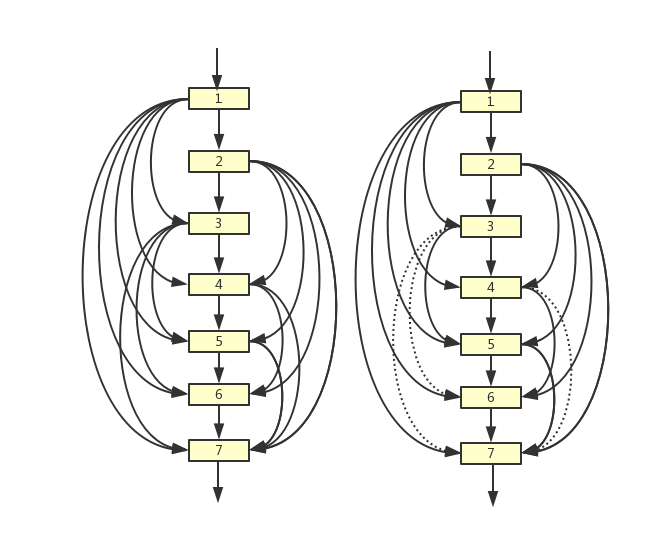}
\caption{Left is DenseNet, input to layers are from all previous layers; right is SparseNet, dotted lines are dropped connections. input to layers are from at most two previous layers.}
\label{fig:example}
\end{figure}

\begin{figure}
\centering
\includegraphics[height=6.5cm]{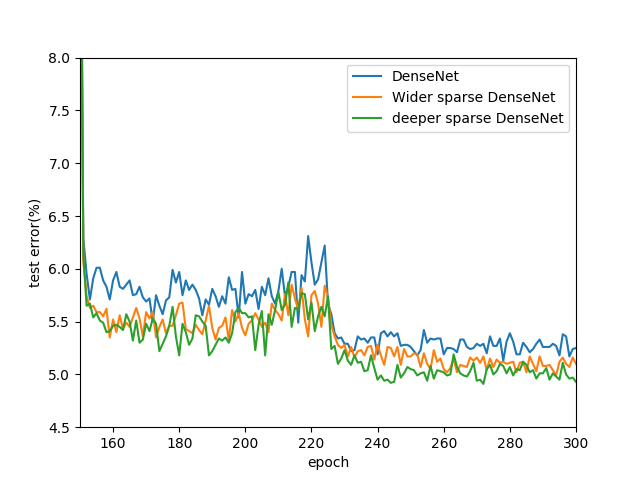}
\caption{Wider Sparse DenseNet and Deeper Sparse DenseNet are networks extended to wider or deeper after drop some middle connections. setup of DenseNet is k(growth rate)=12, layer=40; setup of Wider Sparse DenseNet is k=16, layer=40, path(total connections)=12; setup of Deeper Sparse DenseNet is k=12, path=12, layer=64.}
\label{fig:example}
\end{figure}

 Beside changing networks' depth, width or shortcut connections to boost model's performance, we can also borrow our knowledge about human visual processing mechanism. The most significant feature of human visual system lies in its attention mechanism. When we skim images, we can automatically  focus on important regions, and then devote more attentional resources to those regions. Recently some researchers on computer vision are enlightened by  attention mechanism of human visual system. They designed mechanisms which can firstly select most significant regions in an image(e.g. foreground regions for object segmentation), and then pay more attention to those regions. Attention mechanism has made progresses on various computer vision tasks, such as image classification\cite{wang2017residual},image segmentation\cite{harley2017segmentation}, human pose estimation\cite{chu2017multi} and so on. Recently, Hu\cite{hu2017squeeze} took advantage of attention mechanism from another perspective, he put different amounts of `attentional resources' to different channels of feature maps. To be specific, he increases weights on channels which have informative features and decreases weights on channels which have less useful features. He proposed SE module, which can calibrate feature responses adaptively for different channels in cost of slightly more computation and parameters. The SE module has been proved to be effective for ResNet\cite{he2016deep},Inception\cite{szegedy2015going} and Inception\_ResNet\cite{szegedy2016rethinking}. However, the improvement is ignorable when it applied to our SparseNet. To settle this problem, we present a new attention mechanism. Its structure is shown in Fig 4. It consists of one global average pooling layer and two convolution modules(includes convolution layer, ReLU layer and batch normalization layer). Borrowing idea of shortcut connections, outputs of both global average pooling layer and the first convolution module are taken as input to the second convolution module. And then outputs of the second convolution module are used to calibrate the original network's output.

There are two contributions in our paper:

1) We present an effective way to sparsify DenseNet, which can improve network's performance by simultaneously increasing depth, width and shortcut connections of networks. Besides,

2) we also proposed an attention mechanism, which can further boost network's performance.

\begin{figure}
\centering
\includegraphics[height=6.5cm]{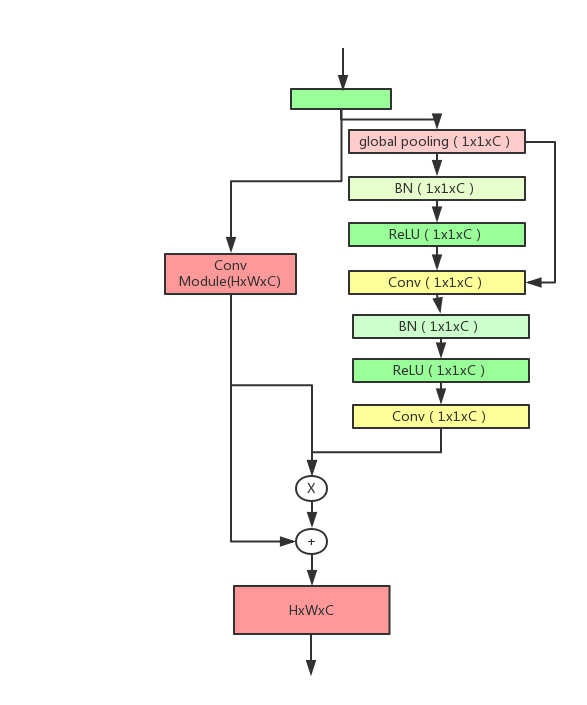}
\caption{SparseNet with attention module}
\label{fig:example}
\end{figure}

\section{Related work}

\subsection{Convolutional neural networks}

Since 2012, neural networks, as a new way of constructing models, have made big steps in various computer vision regions. AlexNet\cite{krizhevsky2012imagenet}, which consists of 8 layers, won the image classification championship of ILSVRC-2012. It reduced error rate on ImageNet dataset from $25.8\%$(best performance in 2011) to $16.4\%$. In 2014, when VGGNet\cite{simonyan2014very} and Inception Net\cite{Szegedy_2015_CVPR} were introduced, depth of networks had been easily extended to 20 layers and the accuracy of image classification also improved a lot. As network goes deeper, simply stacking layers would degrade its performance. To solve the problem, He\cite{he2016deep} introduced ResNet, which learns Residual function ${H(x)-x}$,instead of target function ${H(x)}$ directly. ResNet can be extended to over 100 layers and  the performance can be further improved.

Many researches have been made on ResNet variants. He\cite{he2016identity} changed  the conventional "post-activation" of the weight layers to "pre-activation". To be specific, he put BN layer and ReLU layer before Conv layer. As the result turned out, this identity-mapping change made training easier and thus the performance of networks better. Han\cite{han2016deep} introduced Deep Pyramidal Residual Networks, which increase width gradually layer by layer and rearrange the convolution module. Experiments showed their network architecture has superior generalization ability compared to original ResNet.

 Targ\cite{targ2016resnet} proposed ResNet in ResNet(RiR), which changed  convolution module to a small deep dual-stream architecture. RiR makes network generalize between residual stream which is similar to a residual block and transient stream which is a standard convolutional layer. Huang\cite{huang2016deep}  constructs a very deep ResNet. By randomly dropping some residual modules with probability ${p}$, they can train different shallower subnetworks in the training phase. In the testing phase, they use the whole deep network, whereas recalibrated every residual module with the survival probability $(1-p)$. In this way, ResNet can be expended to over 1200 layers.  Zagoruyko\cite{zagoruyko2016wide} introduced a ResNet variant with wider width and shallower depth, named WRN(Wide Residual Networks), which can improve ResNet's performance further. Huang\cite{huang2017densely} presented DenseNet with layers connected to its all previous layers. With this kind of network design, it not only accomplishes feature reuse, but also alleviates the vanishing-gradient problem.

\subsection{Attention mechanism}
Attention mechanism has achieved many progresses in areas such as machine translation\cite{bahdanau2014neural}. Recently attention mechanism is playing a significant role in various computer vision tasks. Harley\cite{harley2017segmentation}learned weights of pixels in multiple scales using attention mechanism, and calculating the weighted average value as the final result of segmentation. Chu\cite{chu2017multi} improve human pose estimation using multi-context attention module. They use holistic attention model to get global consistency information of human body; while using body part attention module to get detailed information for each human part. Wang\cite{wang2017residual} proposed a residual attention network for image classification, which achieved state-of-art performance on CIFAR dataset. By attention residual learning, they can easily extend their networks to hundreds of layers.  Hu\cite{hu2017squeeze} proposed SENet(Squeeze-Excitation networks), which calibrate weights for different channels by explicitly modeling channel interdependencies. SENet won ILSVRC-2017 image classification championship.

\section{SparseNet}

\subsection{DenseNet}
We represent the input image as $x_0$,output of the  $i^{th}$ layer as $x_i$ and each convolutional module as function $H$. Since input to the  $i^{th}$ layer is outputs of all previous layers. The formula is presented as follows:

$x_i = H([x_0, x_1,…,x_{i-1}])$,

where $[x_0, x_1,…,x_{i-1}]$ is the concatenation of outputs of all previous layers. DenseNet is composed of several dense blocks connected by transition layer. Normally, size of feature map decreased by $\dfrac{1}{4}$ for each block. For example, size of feature map  for the first block is $h\times w$, then $\dfrac{h}{2}\times\dfrac{w}{2}$ for the second block, $\dfrac{h}{4}\times \dfrac{w}{4}$ for the third block. In DenseNet, number of output feature-maps for each convolution module are always the same, which is  denoted by $k$. Thus the output number of the $i^{th}$ layer is $(k_0+(i-1)\times k)$, where $k_0$ is the number input to the first dense block. k was referenced as $growth~rate$.

	As DenseNet goes deeper, number of input feature-maps would become excess very soon.  To settle this problem, the author put $1\times1$ convolution module(as bottleneck layer) before the $3\times3$ convolution module. Thus, the convolutional module has changed from BN+ReLU+$3\times3$Conv to BN+ReLU+$1\times1$Conv+BN+ReLU+$3\times3$Conv.(And the new convolution module is counted as two layers instead of one). The normal setup for output feature-maps of bottleneck layer($1\times1$Conv) is $4k$. Thus, inputs to every $3\times3$ Conv layer is fixed to $4k$. To further improve model compactness, number of feature-maps can  also be reduced in transition layer. The normal setup is number of feature-maps is reduced by $\dfrac{1}{2}$ . This kind of DenseNet is called DenseNet-BC.

\subsection{SparseNet}
We introduce a method to sparsify DenseNet. The basic idea is dropping connections from middle layers and preserving only the farthest and nearest connections. The formula is as followings:

$x_i = H([x_0, x_1,... x_{n/2}, x_{i - n/2} , ... ,x_{i-1}])$,

where $n$ denotes number of connections we will preserve(We call it `path'). As DenseNet does, we also use bottleneck layer and compress the model in the transition layer, the hyperparameters are set the same as that of DenseNet.

Moreover, we also make a structure optimization. In DenseNet, layer number are the same for all dense blocks. However, in our SparseNet, the layer number in each block is increasing. We will talk about the advantages of this arrangement in section 4.6.

\subsection{Attention mechanism}
We proposed an attention mechanism to further boost network's performance. Structure is shown in Figure 3. Suppose the input is $x$, the left part is a convolution module, we denoted the function as $H$. The right part is the attention mechanism module, and denote it as $F$. It consists of one global Pooling layer and two $1\times 1$ convolution modules. The input to the second convolution module is the concatenation of outputs of both global pooling layer and the first convolution module. Then the final result is calculated as $H(x) + H(x) \times F(x)$.

\subsection{Framework}
To summarize, as it is shown in Fig.5, We proposed three networks. (a) is the original DenseNet; (b) is the basic SparseNet(path = $2$, since connections to every layer is at most $2$); (c) is SparseNet-bc, by adding bottleneck layers and reducing number of  feature-maps in transition layer; (d) is SparseNet-abc, by adding attention mechanism on SparseNet-bc. The whole framework is shown in Figure 6.

\begin{figure}
\centering
\includegraphics[height=6.5cm]{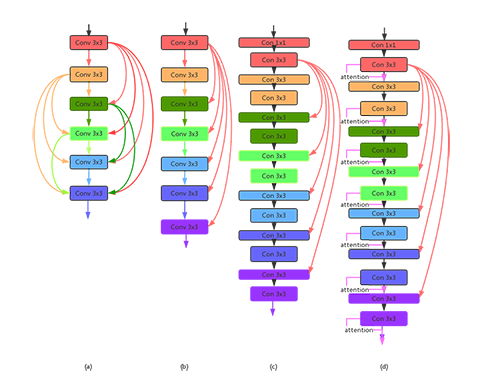}
\caption{a is DenseNet; b is SparseNet(path=2); c is SparseNet-bc; d is SparseNet-abc.}
\label{fig:example}
\end{figure}

\begin{figure}
\centering
\includegraphics[height=1.8cm]{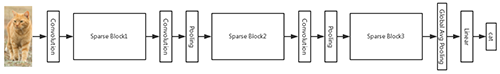}
\caption{the framework of SparseNet for image classification. Between Sparse blocks are transition layer.}
\label{fig:example}
\end{figure}

\subsection{Implementation details}

All our models include three sparse blocks. The layers within each sparse block are increasing. Besides bottleneck layer, all convolutional kernels are $3\times3$. blocks are connected with transition layer, which reduced feature map size by $\dfrac{1}{4}$ and feature map number by $\dfrac{1}{2}$ (feature map number will remain the same for the basic SparseNet).  After the last block, a global pooling layer and a softmax classifier is attached. For each network(SparseNet, SparseNet-bc, SparseNet-abc), we construct three different  sizes of parameters. denoting by V1, V2, V3 and V4. For V1, the layer number for three blocks are 8,12,16; 12,18,24 for V2; 16, 24,32 for V3 and 20,30,40 for V4. other parameters are listed in table 1.

\begin{table}[!]
  \begin{center}
    \caption{setups of networks.}
    \label{tab:headings}
    \begin{tabular}{l|c|c|c|c} 
    \hline
      \textbf{name} & \textbf{\#Params} & \textbf{Depth}& \textbf{Growth rate} & \textbf{Path}\\
            \hline
      SparseNet-V1 & 1.20M & 40 & 16&14\\
      SparseNet-V2 & 5.70M & 68 & 24&21\\
      SparseNet-V3 & 17.5M & 76 & 32 & 28\\
      SparseNet-V4 & 65.7M& 96&50 &35\\
      \hline
      SparseNet-bc-V1 & 0.83M & 76 & 16 & 14\\
      SparseNet-bc-V2 & 3.45M & 132 & 24& 21\\
      SparseNet-bc-V3 & 9.69M & 148 & 32 & 28\\
      SparseNet-bc-V4 & 34.3M &184&50&35\\
      \hline
      SparseNet-abc-V1 & 0.86M & 76 & 16 & 14\\
      SparseNet-abc-V2 & 3.56M & 132 & 24 & 21\\
      SparseNet-abc-V3 & 9.92M & 148 & 32 & 28\\
      SparseNet-abc-V4 & 35.0M &184&50&35\\
      \hline
    \end{tabular}
  \end{center}
\end{table}

\section{Experiments}
\subsection{Datasets}
\subsubsection{CIFAR}

CIFAR\cite{krizhevsky2009learning} are colored images with three channels. Their sizes are $32\times 32$.  CIFAR10 consists of 10 classes and CIFAR100 consists of 100 classes. Both are composed of 50,000 training images and 10,000 test images.

\subsubsection{SVHN}

The Street View House Numbers(SVHN)\cite{netzer2011reading} are also colored images with three channels. Their sizes are $32\times 32$. SVHN includes 73,257 training images, 531,131 additional training images and 26,032 test images. We training our model using all the training images.

\subsection{Training}

All networks are trained using stochastic gradient descent. The weight decay is 0.0001, Nesterov momentum is 0.9 without dampening.  We initialize parameters as He\cite{he2015delving} does. All datasets are augmented with method introduced in Huang\cite{huang2017densely}. For CIFAR, the training epoch is 280. the initial learning rate is 0.1, and decreasing learning rate to 0.01,0.001,0.0002 at epoch 150, 200 and 250. For SVHN, the total epoch is 40, and decreasing to 0.01 and 0.001 at epoch 20 and 30. the batch size of both datasets are 64.

\subsection{Classification Results on CIFAR and SVHN}

Results on datasets of CIFAR and SVHN are shown in table 2. Compared to DenseNet, SparseNet achieves superior performances on all datasets. On CIFAR10, SparseNet decreases error rate from $3.46\%$ to $3.24\%$. On CIFAR100, SparseNet achieves error rate of $16.98\%$, while DenseNet achieved $17.18\%$. On SVHN, SparseNet also achieves lower error rate($1.69\%$ v.s. $1.74\%$ ). Furthermore, SpareNet outperforms the existing state-of-art on CIFAR10 and SVHN. Its error rates are lower than PyramidNet on CIFAR10, which achieved $3.31\%$ and DenseNet on SVHN, which achieved $1.74\%$.

\begin{table}[!]
  \begin{center}
    \caption{Error rates on datasets of CIFAR and SVHN.}
    \label{tab:headings}
    \begin{tabular}{l|c|c|c|c|c} 
    \hline
      \textbf{Networks} & \textbf{\#Params} & \textbf{Depth}& \textbf{CIFAR10} & \textbf{CIFAR100} &\textbf{SVHN}\\
            \hline
      ResNet\cite{he2015delving} & 1.70M & 110 & 6.41&- & -\\
      \hline
      \multirow{2}{*} {ResNet with Stochastic Depth\cite{huang2016deep} }
      & 1.70M & 1.7M& 5.23 &24.58 &{1.75}\\
      & 10.2M & 1202 & 4.91 & - & -\\
      \hline
      \multirow{2}{*}{wide ResNet\cite{zagoruyko2016wide} }
      &11.0M & 16 & 4.81 & 22.07 & - \\
      &36.5M & 28 & 4.17 & 20.50 & -\\
      \hline
      \multirow{2}{*} {pre-activation ResNet\cite{he2016identity} }
       & 1.7M& 164& 5.46& 24.33& -\\
       & 10.2M & 1001 & 4.62 & 22.71 & -\\
      \hline
      \multirow{2}{*} {ResNeXt\cite{xie2017aggregated} }
       & 34.4M& 29& 3.65& 17.77& -\\
       & 68.1M & 29 & 3.58 & 17.33 & -\\
      \hline
      \multirow{2}{*}{FractalNet with Dropout/Drop-path\cite{larsson2017fractalnet}}
      & 38.6M& 21& 5.22& 23.30& 2.01\\
      & 38.6M & 21& 4.60 & 23.73& 1.87\\
      \hline
      PyramidNet($\alpha$=48)\cite{han2016deep} &1.7M &110 & 4.58& 23.12&-\\
      PyramidNet($\alpha$=48)& 28.3M & 110&3.73&18.25&-\\
      PyramidNet($\alpha$=200,bottleneck)&26.0M & 272& 3.31& 16.35& -\\
      \hline
      \multirow{3}{*}{DenseNet-bc\cite{huang2017densely} }
      &0.8M & 100 & 4.51& 22.27 & 1.76\\
      &15.3M & 250 & 3.62 & 17.60 & 1.74\\
      &25.6M & 190 & 3.46& 17.18& - \\
      \hline
      CondenseNet-122\cite{huang2017condensenet} & 0.95M & 122 & 4.48&-&-\\
      CondenseNet-182 & 4.2M & 182 & 3.76 & 18.47&-\\
      \hline
      SparseNet-bc-V1 & 0.83M & 76 & 4.34 & 22.18 & 2.0\\
      SparseNet-bc-V2 & 3.45M & 132& 3.93 & 19.27 &1.85 \\
	  SparseNet-bc-V3 & 9.69M & 148 & 3.56 & 17.75& 1.75\\
      \hline
      SparseNet-abc-V1 & 0.86M & 76 & 4.25& 21.59& 1.98\\
      SparseNet-abc-V2 & 3.56M & 132 & 3.75 & 19.54& 1.89\\
      SparseNet-abc-V3 & 9.92M & 148 & 3.40& 17.53& \textbf{1.69}\\
      SparseNet-abc-V4 & 35.0M & 184 & \textbf{3.24} & 16.98& -\\
      \hline
    \end{tabular}
  \end{center}
\end{table}

 \subsection{Attention mechanism}
 As we can see from table 2, attention mechanism can boost networks' performance for most model sizes(V1, V2 and V3) with only $2\%$ increasing in parameters and $1\%$ increasing in inference time. We also compared our attention mechanism to SE module\cite{hu2017squeeze} on SparseNet-V1. Results are shown in Fig. 7. In the whole training phase, our attention mechanism is always superior to SE module. Besides that, the effect of SE module on SparseNet-V1 is nearly neglectable.

 \begin{figure}
\centering
\includegraphics[height=6.5cm]{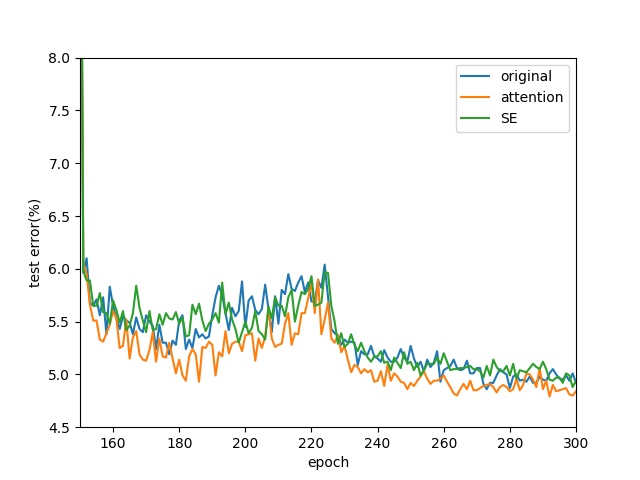}
\caption{original is the SparseNet; attention is SparseNet+attention module; SE is the SparseNet+SE module(the epoch and learning rates are set as DenseNet)}
\label{fig:example}
\end{figure}

\subsection{Parameter Efficiency and Computation Efficiency of SparseNet}
The results in Fig.8 indicate that SparseNet utilizes parameters more efficiently than alternative models. SparseNet-abc-v1 achieves lower test error on CIFAR10 than pre-activation ResNet of 10001 layers, while latter has $10$ times more parameters than the former one. For DenseNet-BC, the best model achieves $3.46\%$, while SparseNet achives lower test errror($3.40\%$) with $\times 2.6$ less parameters. For the recent CondenseNet\cite{huang2017condensenet}, which designed for mobile devices,  Our SparseNet is still more parameter-efficient.

\begin{figure}
\centering
\includegraphics[height=6.5cm]{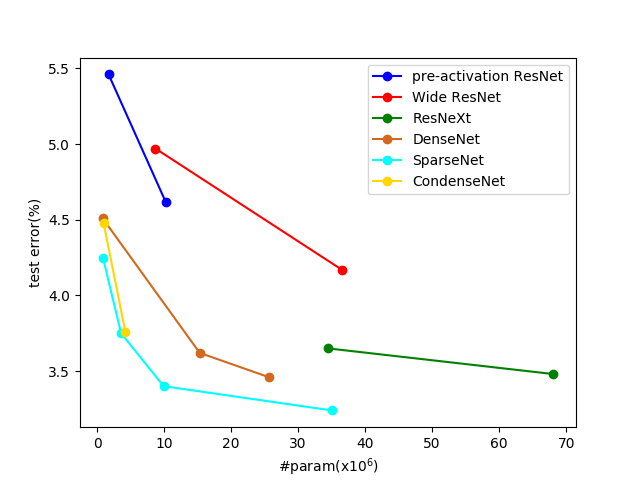}
\caption{Comparison parameter-efficiency on CIFAR10 of different models}
\label{fig:example}
\end{figure}

 To analyze SparseNet's computation, we compared  FLOPs\footnote{computed Conv2D with TensorFlow framwork} (floating-point operations) of pre-activation ResNet, DenseNet and SparseNet.  Results are shown in Fig. 8.  It shows SparseNet is more computation-efficient than the other two models. Compared to the best DenseNet Model,  SparseNet is $\times3.7$ faster than DenseNet.

\begin{figure}
\centering
\includegraphics[height=6.5cm]{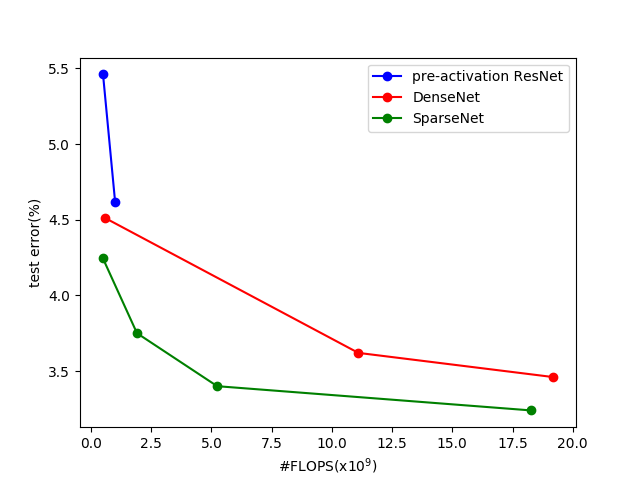}
\caption{Comparison of SparseNet-abc and DenseNet error rate on CIFAR10 as a function of FLOPs.}
\label{fig:example}
\end{figure}

\subsection{Structure optimization}
We also analyzed the effectiveness of our layer arrangement for each sparse block.  We compared two kinds of block arrangements. One is the increasing arrangement: 8-12-16; the other is equal arrangement: 12-12-12.  Results are listed in table 3.  It shows that our increasing arrangement is superior not only in computation but also in accuracy.

\begin{table}[!]
  \begin{center}
    \caption{Results of two block arrangements on CIFAR10.}
    \label{tab:headings}
    \begin{tabular}{l|c|c} 
    \hline
      \textbf{block setup} & \textbf{\#FLOPS} & \textbf{test error}\\
            \hline
      8-12-16 & 578M & 4.95\\
      12-12-12 & 849M & 5.36\\
            \hline

    \end{tabular}
  \end{center}
\end{table}



\section{Discussion}
\subsection{Where to drop connections}
In this section, we experimented different methods of reducing connections. Take SparseNet-V1(path=14) for example, we tried $5$ different ways of dropping connections:

 1)14-0:  only preserving the farthest 14 connections;

 2)10-4:  preserving the farthest 10 connections and nearest 4 connections;

 3)7-7(ours): preserving the farthest 7 connections and nearest 7 connections;

 4) 4-10: preserving the farthest 4 connections and nearest 10 connections;

 5) 0-14: only preserving the nearest 14 connections.

 As we can see from Fig. 10,  different dropping connection methods resulted in different error rates. And our dropping connections method(7-7) achieves best performance. Besides 7-7, 0-14 also achieved comparable performance to our method. One possible explanation is that the method of preserving the nearest 14 connections contains as much information as method of preserving the farthest 7 connections and nearest 7 connections for SparseNet-V1.

\begin{figure}
\centering
\includegraphics[height=6.5cm]{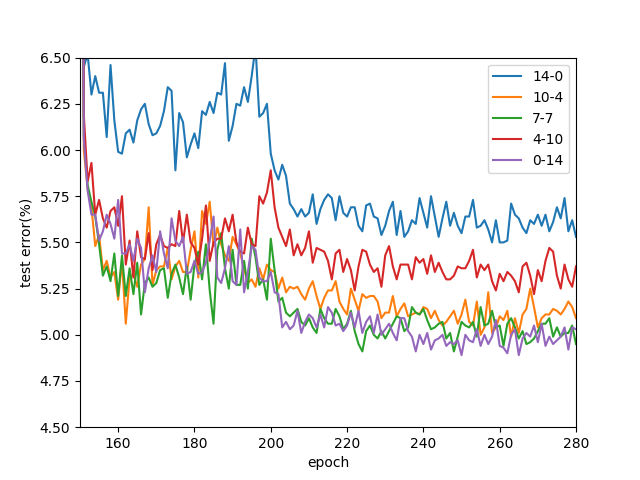}
\caption{different methods of reducing connections in SparseNet-V1.}
\label{fig:example}
\end{figure}

\subsection{How layers, growth rate and path influence network's performance}
We also analyzed how networks' layer, width and shortcut connections influence network's performance. In our experiments, we set up three layers: $28$($8$ layers per block), $52$ layers($16$ layers per block) and $76$ layers($24$ layers per block). We set the range of growth rate(k) to be [6,26]. The parameters of all models are around 1M. So when we set the layer number  and the growth rate, the number of connections(path) is also determined. The results are showed in Fig. 11. We can see that for each layer setup, all test errors are experiencing decreasing first and then increasing, resulting the optimal test error are always in the somewhere middle. For different layer setups, the lowest test error is within layers of $52$, which is between $28$ and $76$. The results showed that none of the three factors shouldn't be  set to be extreme. Only by increasing layers, growth rate and path synchronously,  can SparseNet achieve better performance.

 \begin{figure}
\centering
\includegraphics[height=6.5cm]{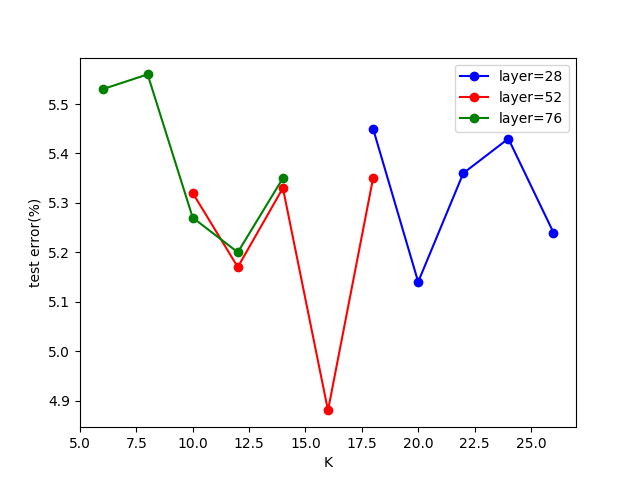}
\caption{Error rates of different setups of layers, growth rate and path on CIFAR10.}
\label{fig:example}
\end{figure}

\section{Conclusion}

In this work, we proposed a method to sparsify DenseNet. After reducing shortcut connections, we can expend the network to deeper and wider. Moreover, we also introduced an attention model, which can boost networks' performance further. Experiments showed that compared to DenseNet, our model achieved comparable performance on datasets of CIFAR and SVHN with much less parameters and much lower computation. Besides, we analyzed several ways of reducing connections and how layers, growth rate and shortcut connections influence networks' performance. In future work, we will apply our models to other computer vision tasks, for example object detection, object segmentation, human pose estimation and so on.

\bibliographystyle{splncs}
\bibliography{egbib}
\end{document}